\documentclass{article}

\usepackage{longcat_style}
\usepackage{adjustbox}
\usepackage[utf8]{inputenc} %
\usepackage[T1]{fontenc}    %
\usepackage{newunicodechar}
\usepackage{hyperref}       %
\usepackage{xcolor}
\usepackage[normalem]{ulem} %
\hypersetup{
    colorlinks=true,      %
    linkcolor=blue,      %
    urlcolor=blue,       %
    citecolor=blue,      %
    linkbordercolor=blue, %
    urlbordercolor=blue,
    citebordercolor=blue,
    pdfborderstyle={/S/U/W 1}, %
}

\usepackage{float}
\usepackage{url}            %
\usepackage{booktabs}       %
\usepackage{amsfonts}       %
\usepackage{nicefrac}       %
\usepackage{microtype}      %
\usepackage{lipsum}		%
\usepackage{graphicx}
\usepackage{natbib}
\usepackage{doi}
\usepackage{amsmath}
\usepackage{amssymb} %
\usepackage{xspace}
\usepackage{enumitem}
\usepackage{multirow}
\usepackage{subcaption} 
\usepackage{makecell}
\usepackage{hyperref, cleveref}
\usepackage{pifont}
\usepackage[inkscapelatex=false]{svg}
\usepackage{subcaption}
\usepackage[most]{tcolorbox} %

\newtcolorbox{summarybox}{
    colback=blue!5!white,    %
    colframe=black!75,       %
    arc=6pt,                 %
    boxrule=1pt,             %
    left=8pt,                %
    right=8pt,               %
    top=8pt,                 %
    bottom=8pt,              %
    enhanced,                %
}

\setlist[itemize]{leftmargin=*}
\setlist[enumerate]{leftmargin=*}
\setlist[description]{leftmargin=*}

\newcommand{\longcat}{LongCat-Flash-Lite\xspace}

\definecolor{midnightgreen}{rgb}{0.0, 0.29, 0.33}

\title{Scaling Embeddings Outperforms Scaling Experts in Language Models}

\author{ Hong Liu, Jiaqi Zhang\footnotemark[1], Chao Wang, Xing Hu, Linkun Lyu, \\
\textbf{Jiaqi Sun, Xurui Yang, Bo Wang, Fengcun Li, Yulei Qian, Lingtong Si, }\\ 
\textbf{Yerui Sun, Rumei Li, Peng Pei\footnotemark[1], Yuchen Xie, Xunliang Cai}  \\
\\
\textbf{Meituan LongCat Team}
}

\renewcommand{\headeright}{\raisebox{-0.2\height}{\includegraphics[height=1.8em]{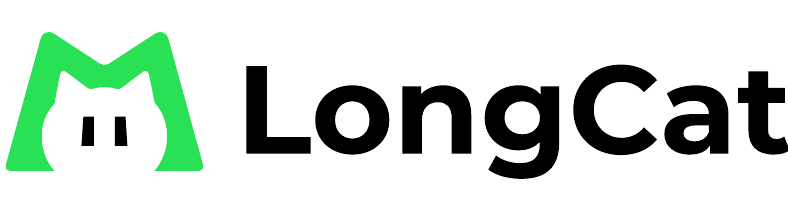}}}
\renewcommand{\shorttitle}{Scaling Embeddings Outperforms Scaling Experts in Language Models}

\newcommand{\method}{N-gram Embedding}

\begin{document}
\maketitle

\footnotetext[1]{Corresponding authors: zhangjiaqi39@meituan.com, peipeng@meituan.com}

\begin{abstract}

While Mixture-of-Experts (MoE) architectures have become the standard for sparsity scaling in large language models, they increasingly face diminishing returns and system-level bottlenecks. In this work, we explore embedding scaling as a potent, orthogonal dimension for scaling sparsity. Through a comprehensive analysis and experiments, we identify specific regimes where embedding scaling achieves a superior Pareto frontier compared to expert scaling. We systematically characterize the critical architectural factors governing this efficacy---ranging from parameter budgeting to the interplay with model width and depth. Moreover, by integrating tailored system optimizations and speculative decoding, we effectively convert this sparsity into tangible inference speedups. Guided by these insights, we introduce \longcat, a 68.5B parameter model with $\sim$3B activated trained from scratch. Despite allocating over 30B parameters to embeddings, \longcat\ not only surpasses parameter-equivalent MoE baselines but also exhibits exceptional competitiveness against existing models of comparable scale, particularly in agentic and coding domains.

\textbf{Hugging Face}: \href{https://huggingface.co/meituan-longcat/LongCat-Flash-Lite}{https://huggingface.co/meituan-longcat/LongCat-Flash-Lite}\\

\end{abstract}

\begin{figure}[h]
    \centering
    \includegraphics[width=0.7\linewidth]{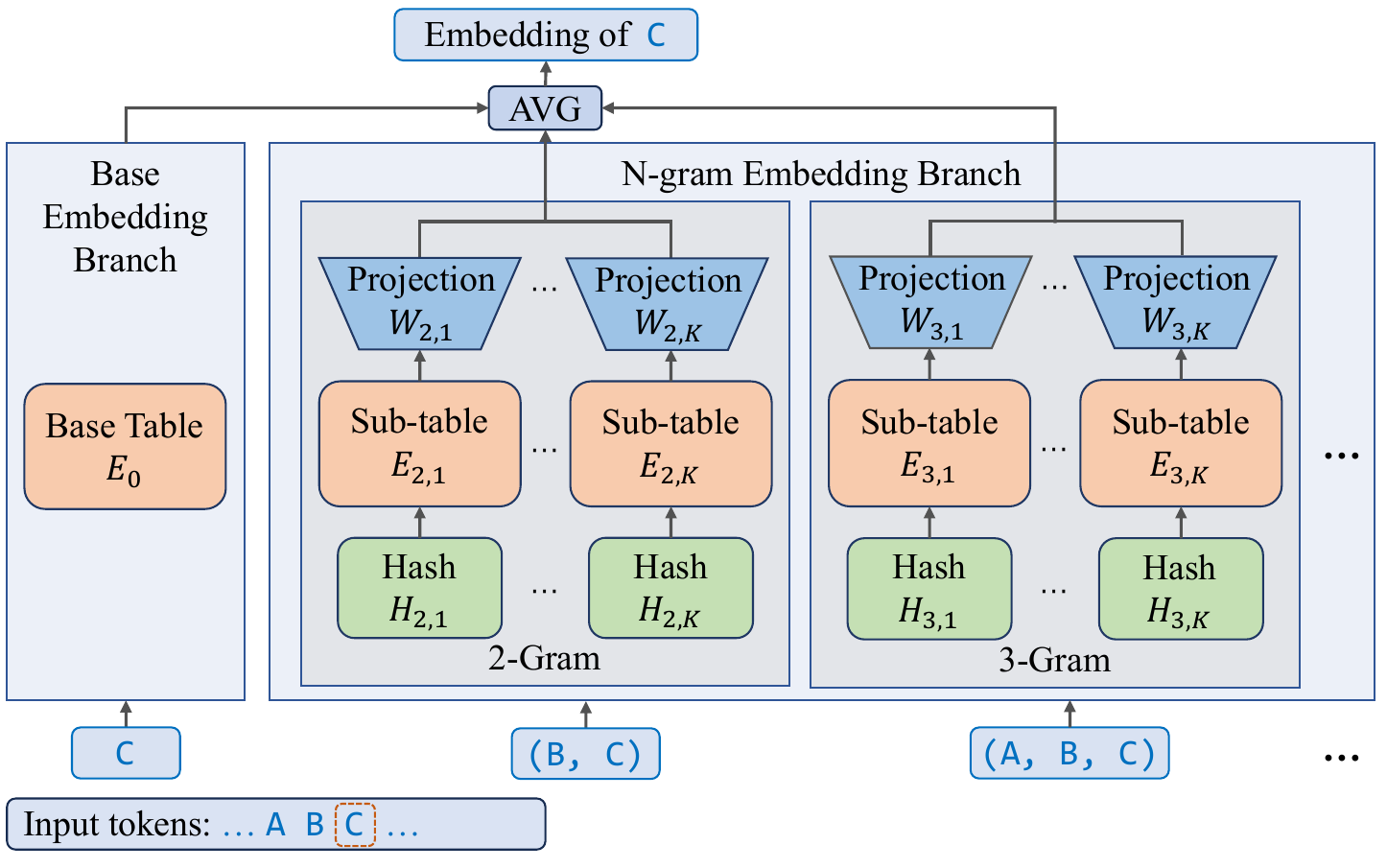}
    \caption{The architecture of a \method~layer~\citep{huang2025overtokenizedtransformervocabularygenerally}. The embedding of each token is augmented by the \method\ branch.}
    \label{fig:overview}
\end{figure}

\clearpage
\tableofcontents
\clearpage

\section{Introduction}
\label{sec:intro}
The Mixture-of-Experts (MoE) architecture has firmly established itself as the dominant paradigm for scaling Large Language Models (LLMs), enabling massive parameter counts while maintaining manageable computational costs \cite{gshard}. By dynamically routing tokens to a subset of experts, models decouple parameter capacity from computational cost, allowing LLMs to scale to trillons of parameters while keeping modest inference latency. However, as the model size and sparsity level increase, the marginal gain in performance diminishes, eventually approaching an efficiency saturation point \citep{apple_scaling}. Furthermore, the practical expansion of experts is constrained by system-level bottlenecks, particularly the escalating communication overhead and memory bandwidth pressure in distributed training. This necessitates the exploration of alternative, orthogonal dimensions for scaling sparse parameters beyond the Feed-Forward Networks (FFNs).

In contrast to MoE, the embedding layer offers an overlooked, inherently sparse dimension with $O(1)$ lookup complexity. This allows for massive parameter expansion without routing overheads—effectively achieving parameter extension without computation explosion. Theoretical foundations for this dimension have been established by scaling laws with vocabulary~\citep{tao2024scaling}, which posit that larger models necessitate proportionally larger vocabularies to maximize computation efficiency. To exploit this potential, diverse strategies have been proposed. One prominent direction is structural expansion, exemplified by Per-Layer Embedding (PLE) \citep{google2025gemma3n, STEM, rosa-plus-github}, which allocates independent embedding parameters to each layer to scale capacity. Another key direction is vocabulary expansion via n-grams to densify information per token. This concept traces back to lookup-table language models \citep{lookuplm} in the RNN era and has recently been advanced in LLMs \citep{clark-etal-2022-canine, huang2025overtokenizedtransformervocabularygenerally, pagnoni-etal-2025-byte, cheng2026conditional}. These approaches collectively highlight the embedding layer as a fertile ground for scaling.

Despite the recent interest in expanding embedding parameters in LLMs, several key challenges remain underexplored. First, the comparative scaling efficiency between expert parameters and embedding parameters is not well understood, leaving the optimal allocation of capacity between these two sparse dimensions ambiguous. Second, the constraints of scaling embeddings are still not systematically characterized: it remains unclear how factors such as the total parameter budget, vocabulary size, initialization schemes, and the trade-offs between model width and depth jointly influence the effectiveness and stability of embedding scaling. Third, while some methods for scaling embeddings have been proposed, it is still unclear which scaling strategy is more effective and efficient under different regimes. Finally, scaling embeddings alters the input/output characteristics of the model during decoding, potentially impacting the overall I/O efficiency, yet its consequences for end-to-end inference performance remain insufficiently analyzed and optimized.

In this technical report, we present a study to address these challenges and establish a robust framework for embedding scaling. Our contributions are as follows: 
\begin{itemize} 
\item \textbf{Comparison of Embedding Scaling vs. Expert Scaling}:
Through comprehensive scaling experiments across diverse scenarios, we identify specific regimes where embedding scaling achieves a superior Pareto frontier compared to increasing expert numbers, offering a high-efficiency alternative for model scaling.
\item \textbf{Impact Analysis of Architectural Factors}:
We establish the complete set of architectural factors determining embedding scaling efficacy, covering the integration timing, parameter budgeting, hash collisions, hyperparameter settings and initialization of embedding, together with the effects of model width and depth. %
Besides, we investigate different methods of scaling embedding and find that \method\ offers the most robust scalability.
\item \textbf{Inference Efficiency and System Optimization}: %
We demonstrate that \method\ largely reduce I/O bottlenecks in MoE layers, particularly when paired with speculative decoding to maximize hardware utilization. Addressing the concomitant embedding overhead, we propose a specialized \textit{N-gram Cache} and synchronized kernels, ensuring that the reduction in active parameters translates directly to lower latency and higher throughput.
\end{itemize}

Based on these findings, we introduce and open-source \longcat, a model trained from scratch with 68.5B total parameters and 2.9B$\sim$4.5B activated parameters depending on the context. Our evaluation demonstrates that \longcat\ not only surpasses a parameter-equivalent MoE baseline—validating the superior efficacy of allocating over 30B parameters to embeddings rather than experts—but also exhibits competitive performance against existing models of similar scale, particularly in agentic and coding tasks.

\section{N-gram Embedding Layer}
\label{sec:ngram-embedding}
To scale the embedding parameters, we adopt the \method\ introduced in \cite{clark-etal-2022-canine, huang2025overtokenizedtransformervocabularygenerally, pagnoni-etal-2025-byte}, which augments the representation of the embedding module by expanding a vocabulary-free n-gram embedding table.
Specifically, for the $i$-th token $t_i$ in a sequence, the augmented embedding $e_i$ is calculated as follows
\begin{equation}
\label{eq:ne-v1}
\begin{aligned}
    &e_i = \frac{1}{N}\big(E_0(t_i) + \sum_{n=2}^N E_n(\mathcal{H}_n({t_{i-n+1}, ..., t_i}))\big), \\
    &\text{where}~ t_{j} = 0 ~\text{if}~ j \leq 0,
\end{aligned}
\end{equation}
where $E_0 \in \mathbb{R}^{V_0 \times D}$ is the original base embedding table with hidden size $D$, $E_n \in \mathbb{R}^{V_n \times D}$ is the expanded embedding table, $N$ denotes the maximum n-gram order and $\mathcal{H}_n$ denotes the hash mapping function.
We use the polynomial rolling hash funciton:
\begin{equation}
\label{eq:hash}
    \begin{aligned}
        \mathcal{H}_n({t_{i-n+1}, ..., t_i}) = (\sum_{j=0}^{n-1} t_{i-j} * V_0^j) \% V_n.
    \end{aligned}
\end{equation}

To further enhance the model's expressive ability and reduce hash collisions, \cite{clark-etal-2022-canine,huang2025overtokenizedtransformervocabularygenerally} decompose each n-gram embedding table into $K$ sub-tables with different vocabulary size. \cite{huang2025overtokenizedtransformervocabularygenerally} further incorporate additional linear projection to map the outputs back to the original embedding space. The final version of \method\ (also referred to as Over-Encoding in \cite{huang2025overtokenizedtransformervocabularygenerally}) is shown in Figure~\ref{fig:overview} and can be written as 
\begin{equation}
\label{eq:ne-v2}
    \begin{aligned}
        e_i &= \frac{1}{(N-1)K+1}\Big(E_0(t_i) + \sum_{n=2}^N \sum_{k=1}^K W_{n,k}E_{n,k}(\mathcal{H}_{n,k}({t_{i-n+1}, ..., t_i}))\Big), \\
    \end{aligned}
\end{equation}
where $E_{n,k} \in \mathbb{R}^{V_{n,k} \times D/((N-1)K)}$ is a sub-table and $W_{n,k} \in \mathbb{R}^{D \times D/((N-1)K)}$ is the linear projection matrix.
By setting the hidden size of sub-tables to be inversely proportional to the number of sub-tables, this design ensures that the parameter count of \method\ remains invariant with respect to $N$ and $K$.

\section{Comparative Analysis of Expert and Embedding Scaling}

This section presents our empirical findings regarding the comparison between scaling embeddings and scaling experts.

\paragraph{Experiment Settings}
We integrate \method\ into the Longcat-Flash architecture \citep{meituanlongcatteam2025longcatflashtechnicalreport} and conduct scaling experiments via from-scratch pre-training across varying activated parameter budgets (280M, 790M, and 1.3B). To rigorously compare scaling strategies, we establish a framework that contrasts scaling via \method\ against scaling experts. 
Specifically, for \method\ scaling, we first train MoE models with varying base sparsity levels ranging from 35\% to 98\% and incrementally incorporate \method\ from specific sparsity levels. %
Crucially, at each sparsity level, the \method\ model is paired with a parameter-equivalent MoE baseline, which attains the same total parameter count by increasing the number of experts.
All models are pre-trained on a corpus of 300B tokens. We evaluate model performance by monitoring training loss and validation loss on two meticulously constructed datasets, covering both Chinese and English.

\subsection{Optimal Timing for N-gram Embedding Integration}

\begin{figure}[t]
    \centering
    \includegraphics[width=0.95\linewidth]{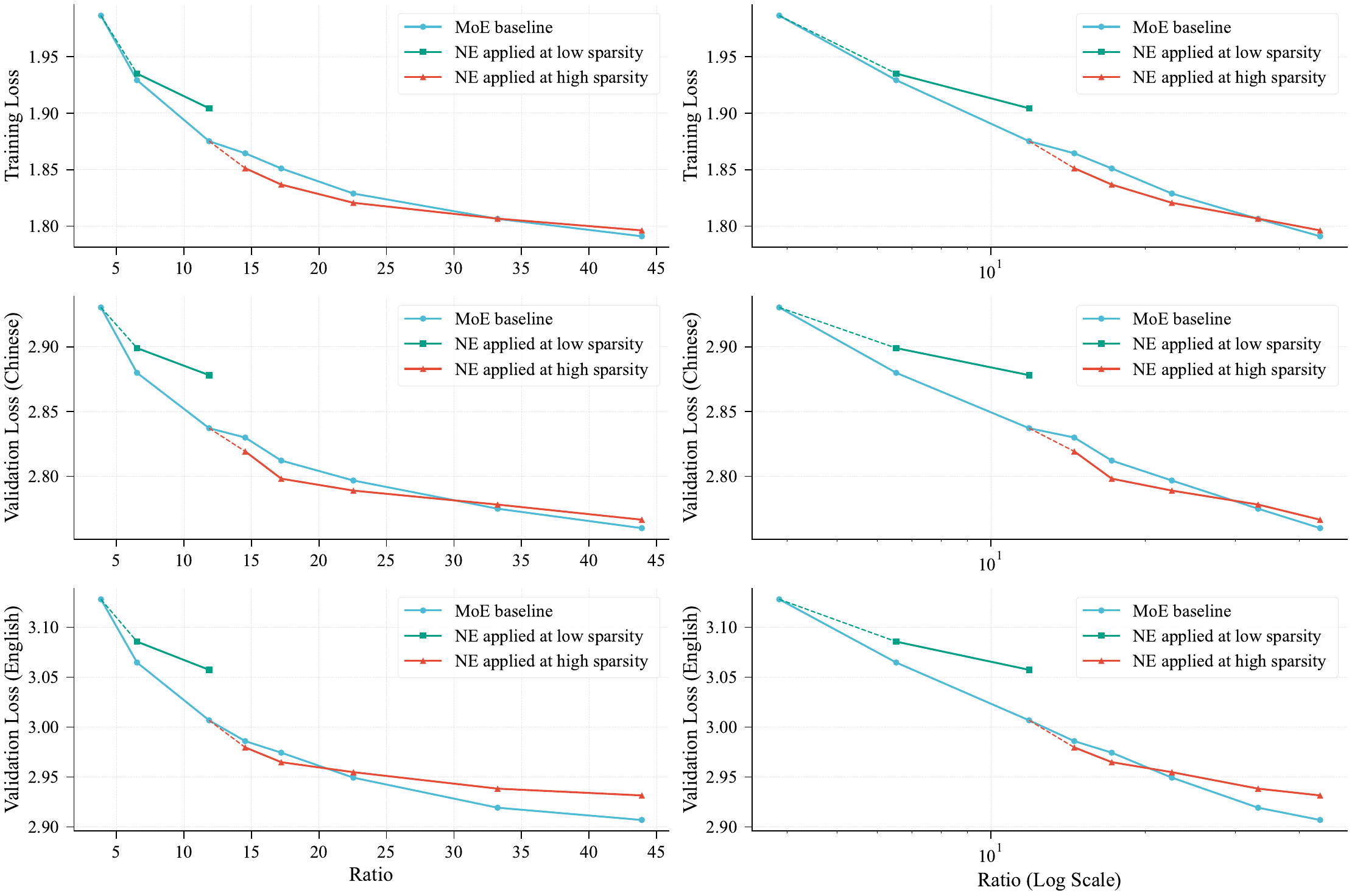}
    \caption{The scaling curve of MoE model and \method\ (NE) model. The horizontal axis is the ratio of total parameters to the activated parameters (280M). The axes of Figures on the right panel is converted to a logarithmic scale. For the two NE curves, we prepend a dashed line to connect the corresponding base MoE model without NE.}
    \label{fig:scaling-200M}
\end{figure}

A pivotal finding is that the scaling dynamics of \method\ diverge markedly depending on the sparsity level of the base model.
Figure~\ref{fig:scaling-200M} presents three distinct scaling trajectories\footnote{We utilize the ratio of total parameters to activated parameters on the x-axis as a proxy for sparsity.}: the standard MoE baseline (blue), \method\ applied to a base model with a low parameter ratio (green), and \method\ applied to a base model with a high parameter ratio (red).
The figure illustrates that the MoE scaling curve adheres to a strict log-linear relationship. This implies that in low-ratio regimes, a marginal increase in the number of experts yields a substantial reduction in loss. Conversely, at higher ratios, achieving an equivalent loss reduction necessitates a significantly larger increase in expert parameters. Consequently, when \method\ is introduced at low parameter ratios, its scaling advantage fails to surpass the gains obtained by simply increasing the number of experts. In contrast, at high sparsity levels, the benefits of \method\ become significantly more pronounced. This observation leads to the following design principle regarding the incorporation of \method.

\begin{summarybox}
\textbf{Summary:} \method\ should be introduced when the number of experts exceeds its ``sweet spot''.
\end{summarybox}

This result indicates that embedding scaling could be a promising scaling dimension orthogonal to expert scaling.

\subsection{Integration Strategy}
\subsubsection{Parameter Budgeting for N-gram Embeddings} 
A closer inspection of Figure~\ref{fig:scaling-200M} reveals a distinct intersection between the blue and red curves: as the parameter ratio increases, the performance advantage of \method\ gradually diminishes and is eventually surpassed by the MoE baseline.
This indicates that when a model allocates an excessive proportion of its parameter budget to \method, its performance becomes inferior to that of parameter-equivalent MoE baselines. This observation aligns with conclusions drawn in the concurrent work Engram \citep{cheng2026conditional}, which posits that the loss follows a U-shaped scaling curve as a function of the \method\ proportion. In Figure~\ref{fig:scaling-200M}, the intersection point lies slightly above a ratio of 20. At this juncture, \method\ parameters constitute approximately 50\% of the total parameter count (given that the base MoE model maintains a ratio of 12). Consequently, we derive a second principle from this phenomenon:
\begin{summarybox}
\textbf{Summary:} Allocate no more than 50\% of the total parameter budget to \method.
\end{summarybox}

\subsubsection{Mitigating Hash Collisions via Vocabulary Sizing}

In the context of \method, hash collisions force a single embedding vector to superimpose the semantics of multiple distinct n-grams. This collision-induced ambiguity impedes learning efficiency and consequently degrades model performance. We identify that selecting an appropriate vocabulary size is critical to mitigating high collision rates. During training, we observed that \method\ exhibits anomalously high hash collision rates at specific vocabulary sizes, particularly for 2-gram hashing. To investigate the underlying mechanics, we conduct a dual analysis focusing on: (1) \textbf{Vocabulary Hit Rate}, defined as the proportion of vocabulary entries activated at least once by the pre-training corpus; and (2) \textbf{Hash Collisions}, which quantifies the loss of unique token representation due to modulo-based indexing.

\begin{figure}[t!]
    \centering
    \begin{subfigure}[b]{0.44\linewidth}
        \centering
        \includegraphics[width=\linewidth]{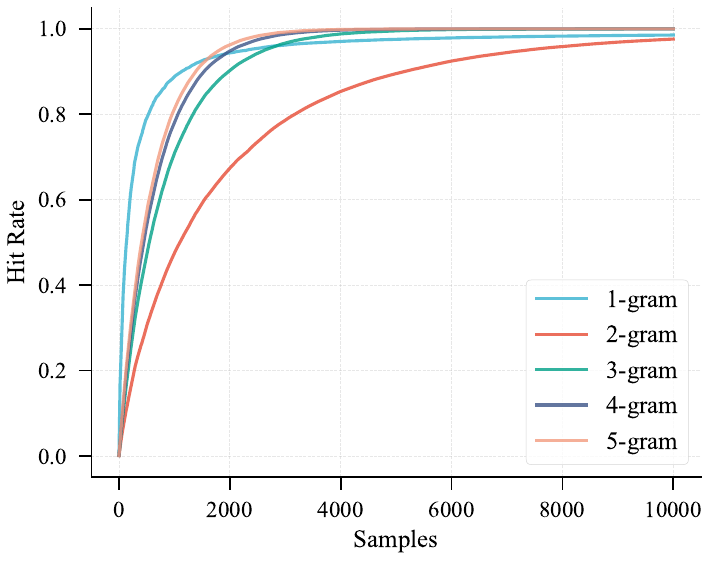}
        \caption{}
        \label{fig:hit-rate}
    \end{subfigure}
    \begin{subfigure}[b]{0.44\linewidth}
        \centering
        \includegraphics[width=\linewidth]{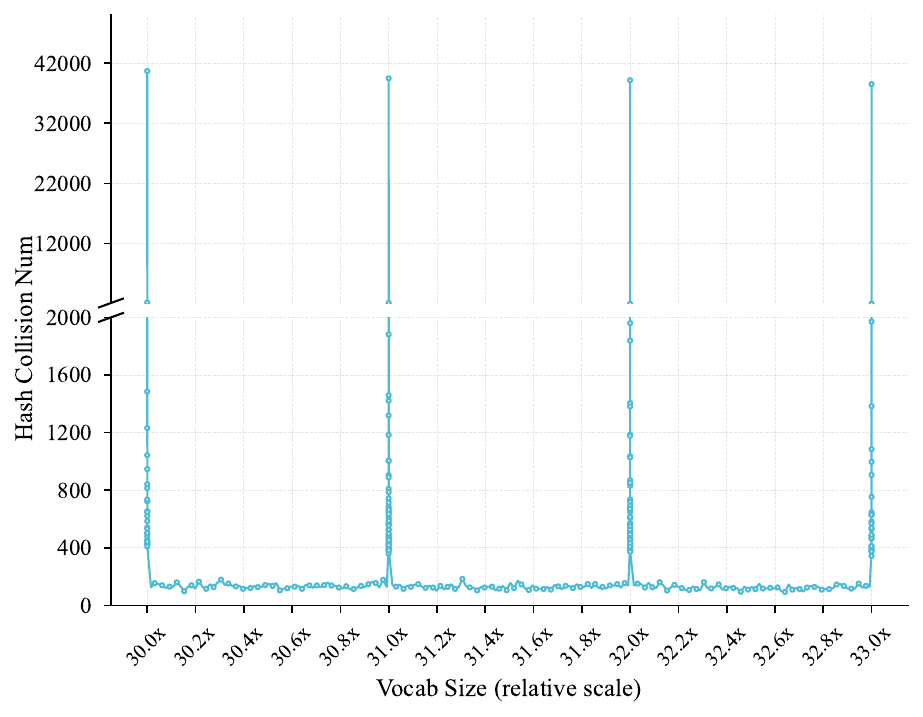}
        \caption{}
        \label{fig:hash-collision}
    \end{subfigure}
    \caption{\textbf{(a)} The vocabulary hit rate of different n-grams. \textbf{(b)} The collision number of 2-gram hashing at different vocabulary size. Sampling points are denser near integer multiples of vocabulary size and sparser elsewhere for clarity.}
    \label{fig:hash-stats}
\end{figure}

For the hit rate analysis, we use an n-gram vocabulary size set to $30\times$ the base vocabulary (128k). For the collision analysis, we sample a range of vocabulary sizes between $30\times$ and $33\times$ the base vocabulary size, computing n-gram collision counts over 100 training sequences for each configuration. The results are detailed in Figure~\ref{fig:hash-stats}.

Figure~\ref{fig:hit-rate} illustrates that 2-gram hashing exhibits a gradual increase in hit rate, whereas higher-order n-gram hashing rapidly converges toward a hit rate of 1.0. 
Independent of the hit rate trends, we observe in Figure~\ref{fig:hash-collision} that 2-gram hashing collision numbers display a strong, non-linear correlation with vocabulary size. A salient pattern emerges: collision counts spike noticeably when the vocabulary size approaches an integer multiple of the base vocabulary size. This phenomenon persists regardless of whether the n-gram vocabulary size is a prime number. Synthesizing these observations, Figure~\ref{fig:hash-collision} motivates an additional design principle for configuring \method:

\begin{summarybox}
\textbf{Summary:}  The vocabulary size of \method\ should significantly deviates from integer multiples of the base vocabulary size to prevent Hash collisions.
\end{summarybox}

\subsubsection{Sensitivity Analysis of Hyperparameters}
We now examine the sensitivity of model performance to the internal configurations of \method, namely the n-gram order $N$ and the number of sub-tables $K$.

Regarding the n-gram order $N$ defined in Section~\ref{sec:ngram-embedding}, increasing $N$ enables \method\ to capture richer contextual semantics, theoretically yielding embedding vectors with enhanced representational capacity. However, this also creates an extremely sparse distribution over the n-gram vocabulary, as high-order n-grams appear infrequently. This sparsity significantly exacerbates the challenge of learning effective embeddings. 

Regarding the number of sub-tables $K$, this parameter governs the number of distinct hash functions applied to each n-gram, thereby substantially mitigating the probability of hash collisions. Nevertheless, empirical evidence suggests that increasing $K$ beyond a certain threshold yields diminishing returns.

\begin{figure}[t]
    \centering
    \includegraphics[width=0.9\linewidth]{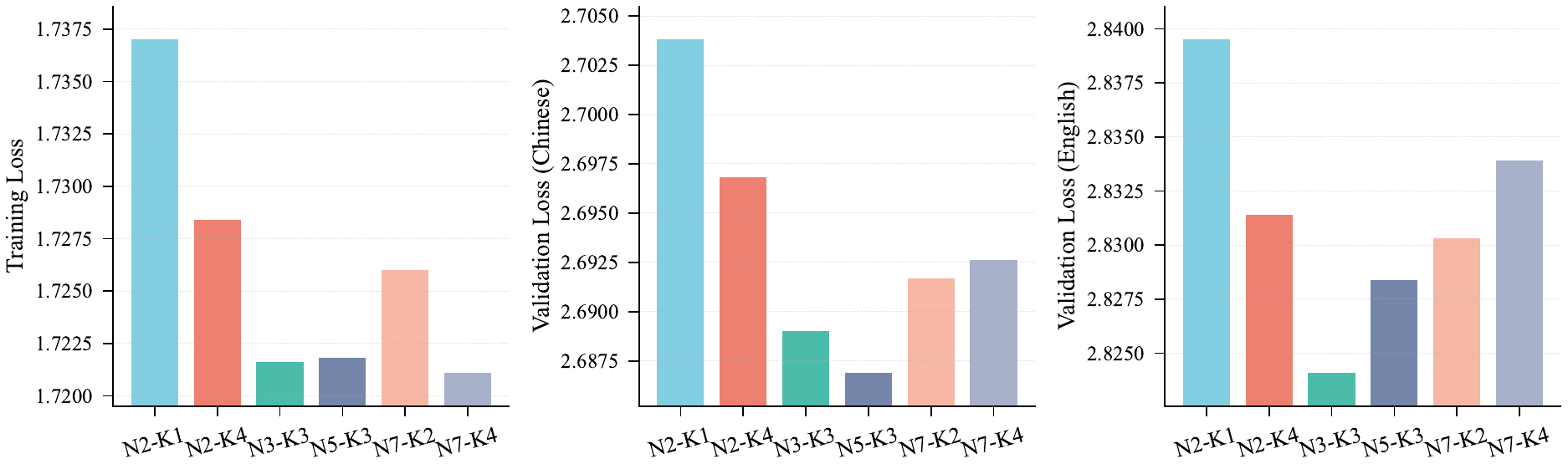}
    \caption{Comparison of training and validation loss under different combinations of $N$ and $K$.}
    \label{fig:loss-nk}
\end{figure}

Utilizing the 790M activated-parameter model (corresponding to the initial data point on the red curve in Figure~\ref{fig:scaling-500M}), we conduct ablation studies across various combinations of $N$ and $K$\footnote{For large $N$, numerical overflow during hash computation can be circumvented by applying the modulus operation prior to exponentiation.}. The results are summarized in Figure~\ref{fig:loss-nk}. It is evident that when both $N$ and $K$ are set to their minimal values ($N = 2$ and $K = 1$), the model exhibits notably inferior performance. Conversely, for $N \ge 3$ and $K \ge 2$, the performance variance across different configurations becomes relatively small, indicating that the model is robust to hyperparameter selection within this regime. Empirically, we observe that setting $N$ in the range of 3 to 5 consistently yields near-optimal performance.

\subsubsection{Embedding Amplification for Effective Training}

\begin{figure}[t]
    \centering
    \includegraphics[width=0.9\linewidth]{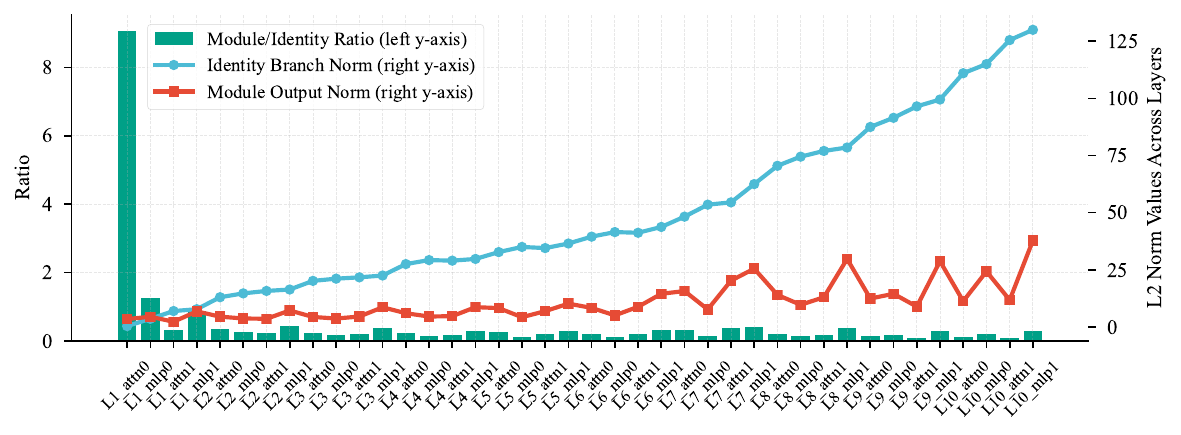}
    \caption{Layer-wise analysis of L2 norms for module outputs versus their corresponding identity branches, alongside the ratio of these norms. Each shortcut layer comprises two sub-layers, denoted by suffixes 0 and 1.}
    \label{fig:norm-per-layer}
\end{figure}

In our preliminary experiments, we observed that suboptimal initialization of the embedding module can severely impede the efficacy of \method, preventing it from realizing its full potential. To validate this hypothesis, we revisited an early vanilla experiment configured as per Figure~\ref{fig:scaling-200M}, but without any specific adjustments to the embedding module. After pre-training on 300B tokens, we compute the L2 norms of each module’s output and its corresponding residual branch (identity path) across all layers, plotting the norms and their ratios in Figure~\ref{fig:norm-per-layer}.

Figure~\ref{fig:norm-per-layer} exposes a critical disparity: the L2 norm of the first attention module's output is an order of magnitude larger (approximately $10\times$) than that of the corresponding identity branch, which essentially represents the output of the embedding module. This indicates that upon summation, the attention output dominates the residual stream, effectively "drowning out" the embedding signal. Although standard initialization of sub-tables and projection matrices in \method\ ensures that initial output norms match the baseline, this signal suppression phenomenon exacerbates significantly once training progress, leading to substantial performance degradation in \method\ models.

To mitigate this issue, we explore two strategies:
\begin{itemize}
    \item \textbf{Scaling Factor:} Introducing a scaling factor (typically $\sqrt{D}$) to the embedding output to ensure a sufficient contribution to the forward pass.
    \item \textbf{Normalization:} Applying LayerNorm to the embedding output prior to merging with the residual branch. This similarly amplifies the embedding contribution, as LayerNorm enforces unit variance during the early stages of training.
\end{itemize}

Both techniques were originally proposed in \cite{takase2025spikemorestabilizingpretraining} with the primary objective of increasing residual branch variance to bound backward gradients and stabilize training. In our context, while we observed no significant impact on training stability, these methods---collectively termed \textbf{Embedding Amplification}---substantially enhance the performance of \method. 
In our experiments, applying Embedding Amplification yields superior performance compared to the vanilla baseline, with a consistent reduction of 0.02 in both the training loss and the two validation losses.

\subsection{Scaling Properties across Model Width and Depth}
\subsubsection{Enhanced Advantage in Wider Models} 
This section investigates how the efficacy of N-gram embedding scaling evolves with increasing model width.

\begin{figure}[t!]
\centering
\begin{subfigure}[b]{0.49\linewidth}
\centering
\includegraphics[width=\linewidth]{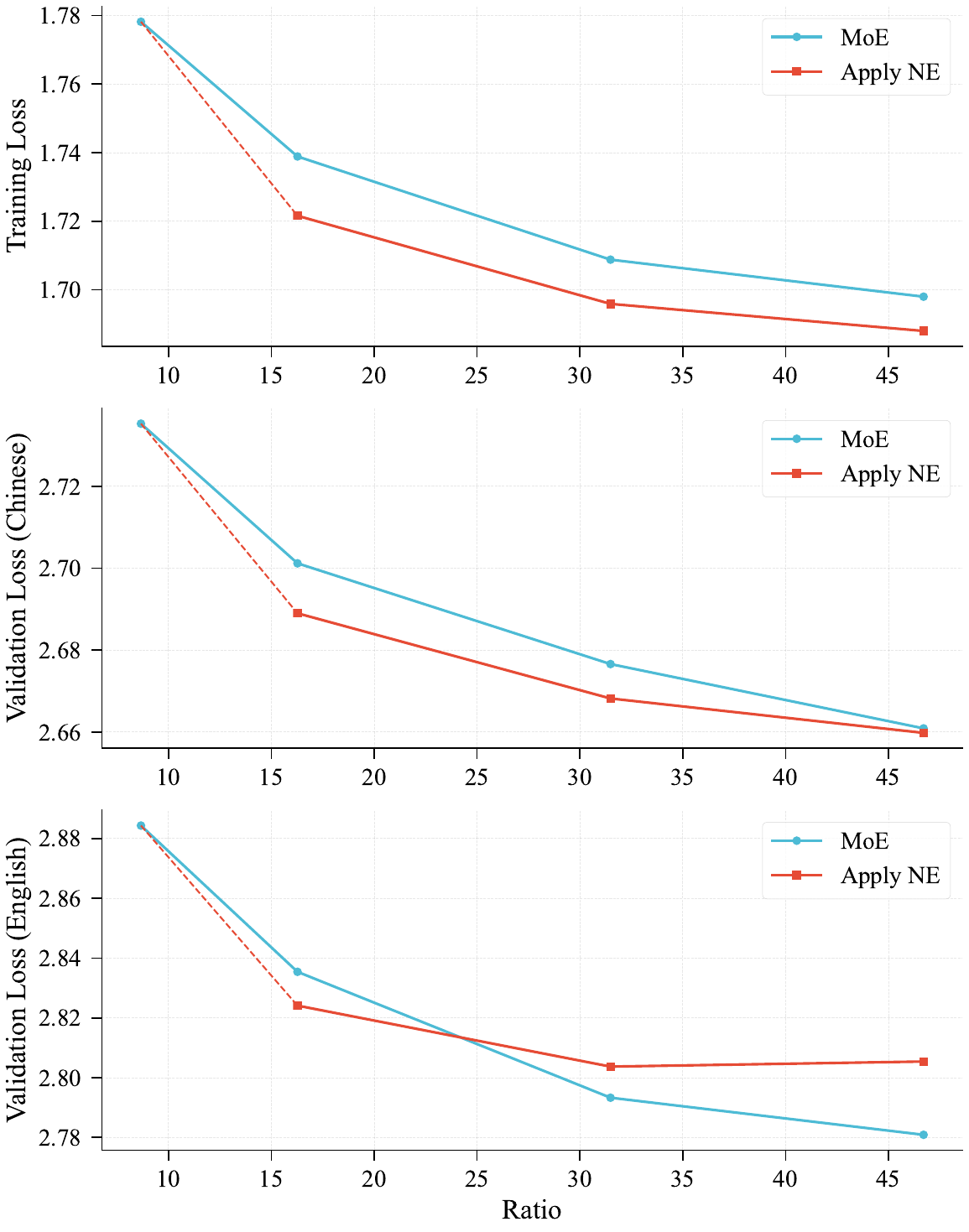}
\caption{}
\label{fig:scaling-500M}
\end{subfigure}
\hfill
\begin{subfigure}[b]{0.49\linewidth}
\centering
\includegraphics[width=\linewidth]{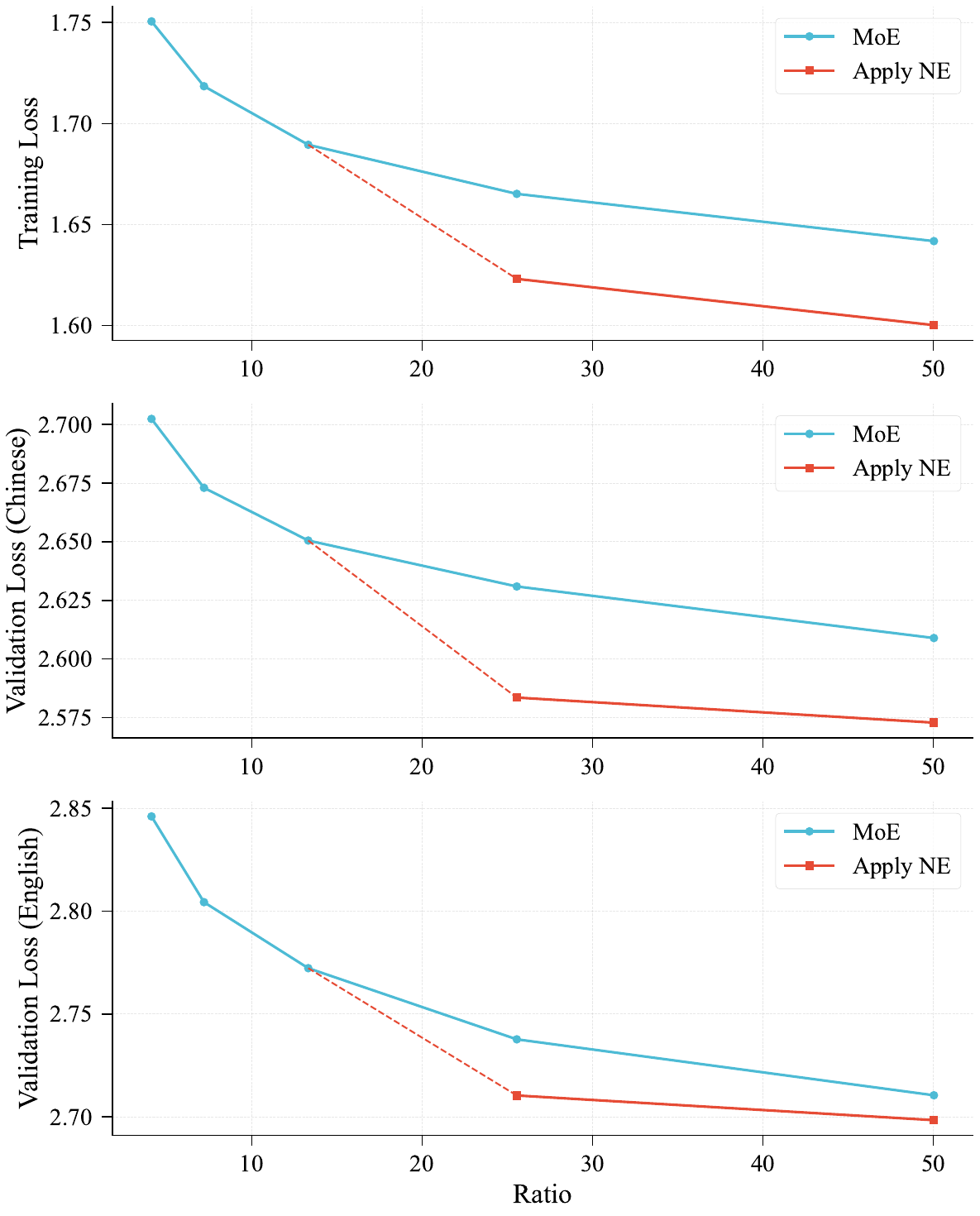}
\caption{}
\label{fig:scaling-1B}
\end{subfigure}
\caption{\textbf{(a)} Scaling curve at 790M activation size. \textbf{(b)} Scaling curve at 1.3B activation size.}
\label{fig:scaling-large}
\end{figure}

We conduct a series of scaling experiments at two larger activation scales (790M and 1.3B parameters), with model depth held constant (10 shortcut layers) while only the width (hidden size and module dimensions) varies.
Figure~\ref{fig:scaling-large} presents the resulting scaling curves. Our analysis reveals two key trends:
\begin{itemize}[leftmargin=*]
\item When incorporated at an appropriate ratio, \method\ consistently yields a lower loss compared to a parameter-equivalent MoE baseline. This advantage gradually diminishes as the proportion of \method\ parameters increases, mirroring the behavior observed in Figure~\ref{fig:scaling-200M}.
\item Crucially, the intersection point between the \method\ curve and the MoE curve systematically shifts towards higher total-to-activated parameter ratios as the model width (activation size) increases. Specifically, for a 280M activation size, \method\ consistently underperforms its MoE counterpart once the ratio exceeds 30. At 790M, \method\ only underperforms on the English validation set at this ratio, while maintaining an advantage on all other metrics. Notably, at a 1.3B activation size, \method\ retains a clear advantage even at ratios as high as 50.
\end{itemize}

These findings demonstrate that wider models allow for a significantly expanded window of opportunity to leverage \method\ effectively. Consequently, Figure~\ref{fig:scaling-large} leads to the following conclusion: for a fixed number of layers,
\begin{summarybox}
\textbf{Summary:} Increasing model width confers a greater advantage to \method.
\end{summarybox}

\subsubsection{Diminishing Returns in Deeper Models} 

We now investigate the impact of model depth on the efficacy of \method\ scaling. For pre-normalization architectures, the contribution of \method\ through the identity connection (residual branch) inherently diminishes as network depth increases, as the signal propagating through skip connections carries less direct information from earlier layers (also shown in Figure~\ref{fig:norm-per-layer}).

To probe this hypothesis, we conducted scaling experiments using deeper architectures, building upon our 1.3B activated parameter configuration. Specifically, we trained models with 20 and 40 layers while meticulously maintaining a consistent relative proportion of \method\ parameters (~50\% of the total parameters) across all tested depths.

\begin{figure}[t!]
    \centering
    \begin{subfigure}[b]{0.45\linewidth}
        \centering
        \includegraphics[width=\linewidth]{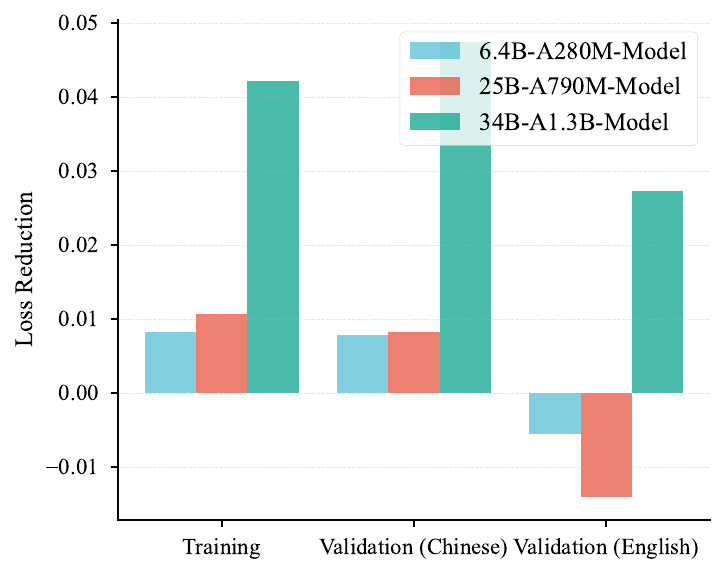}
        \caption{}
        \label{fig:loss-diff-width}
    \end{subfigure}
    \hfill
    \begin{subfigure}[b]{0.45\linewidth}
        \centering
        \includegraphics[width=\linewidth]{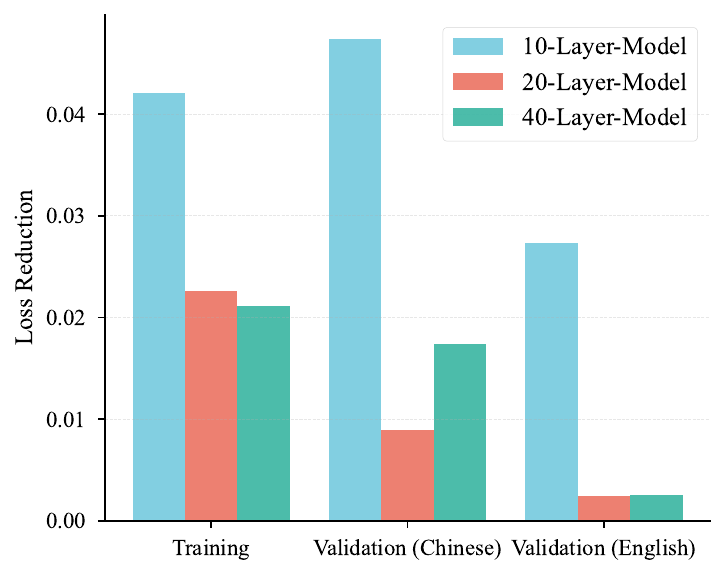}
        \caption{}
        \label{fig:loss-diff-depth}
    \end{subfigure}
    \caption{Loss reduction of models with \method\ compared to the baseline across different \textbf{(a)} model width and \textbf{(b)} model depth.}
    \label{fig:loss-diff}
\end{figure}

Figure~\ref{fig:loss-diff-depth} presents a clear comparison of the performance gap between \method\ and the MoE baseline across these varying depths. A striking observation emerges: as model depth surpasses 20 layers, the performance advantage of \method\ over the baseline experiences a pronounced contraction.
This trend stands in contrast to the effect of increasing model width, as illustrated in Figure~\ref{fig:loss-diff-width}, where the performance gap demonstrably widens.

\begin{summarybox}
\textbf{Summary:} Increasing model depth diminishes the relative advantage of \method.
\end{summarybox}

Note that the majority of current practical language models typically operate below 40 shortcut layers (equivalent to 80 conventional layers). Given our finding that increased width consistently amplifies \method's advantage, and its robust performance even at 40 layers, scaling n-gram embeddings up within these common architectural depths is likely to yield even greater performance gains.

\section{Efficient Inference}
\label{sec:inference}

\subsection{Reduction of MoE Activation Parameters}

The N-gram Embedding mechanism effectively redistributes parameters from the MoE layers to the embedding space. This architectural transformation maintains the total model parameters while reducing the number of activated parameters within MoE layers—particularly advantageous in memory I/O-bound decoding scenarios with large token counts. Moreover, the increased size of the embedding layer does not penalize latency, as the computational cost of embedding lookups scales with the number of input tokens rather than the total number of embedding parameters.

\begin{figure}[t!]
    \centering
    \begin{subfigure}[b]{0.4\linewidth}
        \centering
        \includegraphics[width=\linewidth]{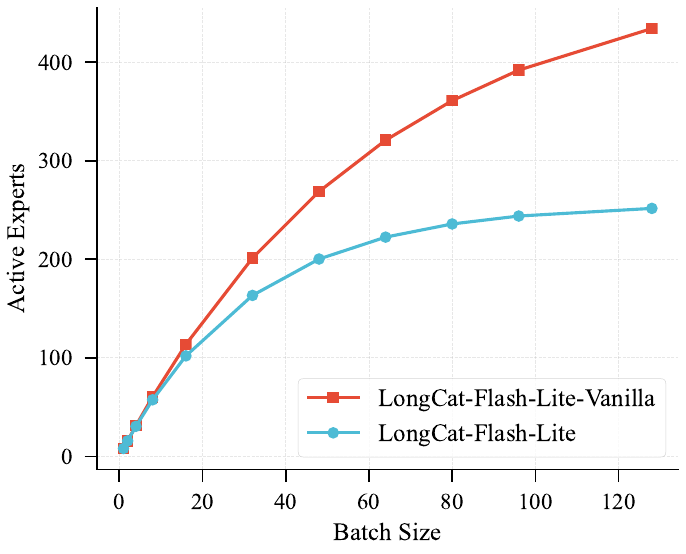}
        \caption{}
        \label{fig:activation-rate}
    \end{subfigure}
    \begin{subfigure}[b]{0.4\linewidth}
        \centering
        \includegraphics[width=\linewidth]{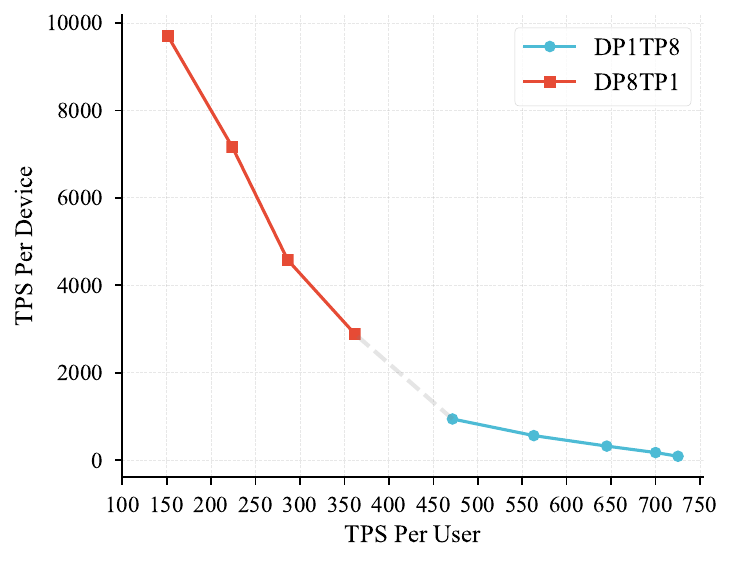}
        \caption{}
        \label{fig:decode-performance}
    \end{subfigure}
    \caption{\textbf{(a)} Number of activated experts in \longcat versus \longcat-Vanilla across varying batch sizes. \textbf{(b)} \longcat decoding performance on 8xH800-80G with ISL=4K and OSL=1K. The middle segment is for visual continuity. The model information of \longcat is described in Section~\ref{sec:longcat}.}
    \label{fig:infer}
\end{figure}

To fully capitalize on the efficiency gains from reduced active parameters, it is crucial to maximize hardware utilization through a large batch size (as shown in Figure~\ref{fig:activation-rate}). This requirement creates a natural synergy with speculative decoding. Multi-step speculative decoding effectively expands the ``effective batch size'', thereby converting the theoretical advantage of parameter sparsity into tangible inference speedups.

\subsection{Optimized Embedding Lookup}
Although reallocating parameters from experts to \method\ effectively reduces memory I/O for MoE layers, it introduces additional overhead in terms of I/O, computation, and communication compared to a standard embedding layer. Minimizing the latency and resource consumption of \method\ is therefore critical for overall system efficiency. Furthermore, the dynamic and complex scheduling mechanisms inherent in modern inference frameworks make it difficult to pre-determine the exact token sequences for the forward pass, which complicates the optimization of N-gram embedding lookups.

To address these challenges, we introduce the \textit{N-gram Cache}, a specialized caching mechanism inspired by the design principles of the KV cache. We implement custom CUDA kernels to manage N-gram IDs directly on the device, facilitating low-overhead synchronization with the intricate scheduling logic of various inference optimization techniques. This design significantly enhances the computational efficiency of N-gram embeddings.

In speculative decoding scenarios, where the draft model typically operates with fewer layers and substantially lower latency, the overhead of \method\ becomes relatively more pronounced. To mitigate this, we propose two complementary optimization strategies: (1) employing a conventional embedding layer for the draft model to bypass the more computationally expensive n-gram lookup; and (2) caching n-gram embeddings during the drafting phase to eliminate redundant computations during the subsequent verification step. These optimizations collectively reduce latency and improve throughput in speculative inference settings.

\subsection{Rethinking \method\ Optimization: The Role of Speculative Decoding}
Beyond hardware efficiency, we posit that the \method\ structure inherently encodes rich local context and token co-occurrence information, offering unexplored synergies with speculative decoding. We identify two promising directions where the semantic richness of \method\ could potentially be leveraged to further accelerate inference.

\textbf{\method\ based drafting}: Since the \method\ aggregates information from the preceding N-1 tokens, it implicitly captures short-range dependencies. We are currently exploring architectures to repurpose the N-gram embedding as an ultra-fast draft model. While a primary candidate involves attaching a lightweight linear projection directly to the \method\ outputs, we are investigating a broader design space to fully exploit the captured local context for efficient token prediction.

\textbf{Early rejection}: The \method\ representation could also serve as a semantic consistency check (or confidence estimator) for tokens generated by external draft models. Draft tokens that result in low-probability match under \method\ might be "early-rejected" before entering the expensive verification phase of the target model. Theoretically, this pruning strategy would reduce the workload of the verification step, offering a pathway to further optimize end-to-end latency.

\section{Integration with Per-Layer Embedding}
As mentioned in Section~\ref{sec:intro}, Per-Layer Embedding (PLE) is another way to scale parameters by allocating embedding parameters across layers. This section provides a direct comparison between \method\ and PLE, and introduces an attempt to integrate both approaches.

\subsection{Per-Layer Embedding}
PLE is applied in \cite{gemma3n-doc} and further studied in \cite{STEM}. PLE directly substitutes the output of up-projection matrix in the SwiGLU module with the embedding output, which is the most efficient method for injecting embedding information in our experiments. Let $x^{(l)}_i$ be the $i$-th input vector of the FFN module in layer $l$, the FFN output with PLE can be formalized as follows
\begin{equation}
    \begin{aligned}
        \text{FFN}^{(l)}(x_i) = W_d^{(l)}(\text{SiLU}(W^{(l)}_gx^{(l)}_i) \odot E_0^{(l)}(t_i))
    \end{aligned}
\end{equation}
where $W_d^{(l)}$ and $W^{(l)}_g$ denote the down-projection and gate-projection matrices of layer $l$ respectively, and $E_0^{(l)}$ is the embedding table of layer $l$, with identical shape to the base embedding table in Eq.~\ref{eq:ne-v1}.

\subsection{Per-Layer N-gram Embedding}
Building upon PLE, we propose Per-Layer N-gram Embedding (PLNE), a novel extension that replaces the base embedding outputs with \method\ outputs at each layer, thereby enabling more flexible and targeted parameter scaling within the MoE framework. PLNE can be written as
\begin{equation}
        \text{FFN}^{(l)}(x_i) = W_d^{(l)}(\text{SiLU}(W^{(l)}_gx^{(l)}_i) \odot e^{(l)}_i)
\end{equation}
where $e^{(l)}_i$ is computed according to Eq.~\ref{eq:ne-v2}, with layer-specific embedding table and projection matrix.

\subsection{Empirical Comparison}
\begin{figure}[t]
    \centering
    \includegraphics[width=0.9\linewidth]{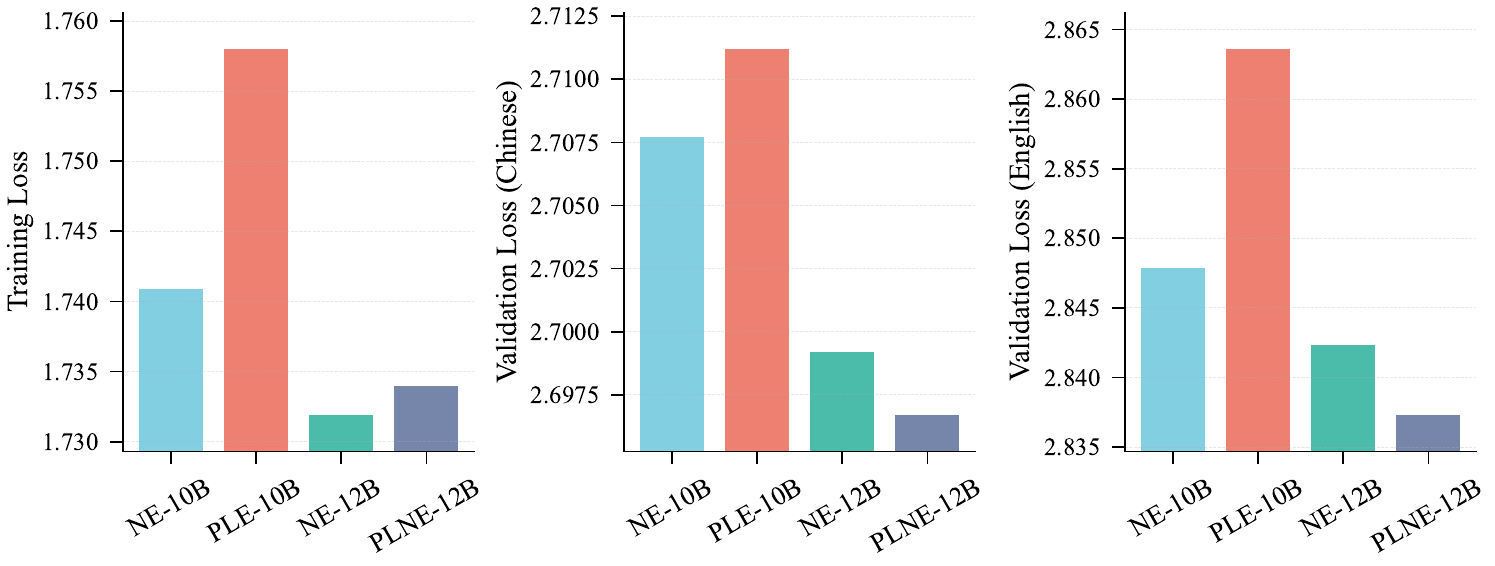}
    \caption{Loss comparison of \method\ (NE), PLE, and PLNE under the 790M activated-parameter setting. Note that PLE and PLNE are compared against NE at two distinct parameter scales.}
    \label{fig:loss-ple}
\end{figure}

For both PLE and PLNE, embedding information is injected exclusively into the MLP within the dense sub-layer of each shortcut layer. 
Since each PLNE layer incorporates an n-gram vocabulary in addition to the base vocabulary, it introduces a larger number of parameters per layer compared to PLE. To avoid confounding factors related to layer positioning, we do not directly compare between PLE and PLNE under equivalent total parameter counts. 
Instead, we evaluate PLE and PLNE against their respective parameter-equivalent \method\ (NE) baselines, as illustrated in Figure~\ref{fig:loss-ple}.

Figure~\ref{fig:loss-ple} reveals that PLE underperforms relative to \method, whereas PLNE yields marginal improvements over NE. We attribute the former to the superior learning efficiency of \method\ compared to standard embeddings.
Consequently, we focused our scaling analysis on PLNE.
However, in subsequent experiments involving increased model width or depth, PLNE failed to exhibit a consistent advantage, performing on par with NE in most scenarios. Given that PLNE inherently increases activated parameters (due to the addition of a substantial projection matrix in each layer), we opted not to adopt PLNE for our larger-scale experiments. Nonetheless, this approach merits further investigation—specifically regarding the optimal allocation of embedding parameters across layers, such as determining whether to concentrate them in a few specific layers or distribute them uniformly throughout the network.

\section{\longcat}
\label{sec:longcat}
Leveraging the insights from our previous analysis, we introduce \longcat, a model trained from scratch with integrated \method. \longcat undergoes a complete pipeline of pre-training, mid-training, and supervised finetuning, and demonstrates highly competitive performance for its scale.

\subsection{Model Information} 

\paragraph{Architecture} \longcat adopts the same architecture as Longcat-Flash \citep{meituanlongcatteam2025longcatflashtechnicalreport}, with a total of 14 shortcut layers. It has 68.5 billion total parameters and dynamically activates between 2.9B and 4.5B parameters per token due to the zero-experts. In each shortcut layer, the MoE module consists of 256 FFN experts and 128 zero-experts, and each token selects 12 experts. For embedding module, \longcat includes 31.4B \method\ parameters, accounting for 46\% of the total.

\paragraph{Training Data} \longcat follows the same data recipe with LongCat-Flash-Chat \citep{meituanlongcatteam2025longcatflashtechnicalreport}. It is first pre-trained on 11T tokens with a sequence length of 8k, followed by 1.5T tokens of mid-training during which the sequence length is extended to 128k, and is finally trained on SFT data. To support extended context, we implement YARN~\citep{Peng2023YaRNEC} during the 32k sequence length training stage, enabling \longcat to handle sequences up to 256k tokens.

\paragraph{Baseline without \method} We train an MoE baseline with exactly the same parameters as \longcat (referred to as \longcat-Vanilla) by converting all \method\ parameters into additional experts. Both models undergo identical training strategy and data recipe.

\subsection{Base Model Evaluation} 
\begin{figure}
    \centering
    \includegraphics[width=0.6\linewidth]{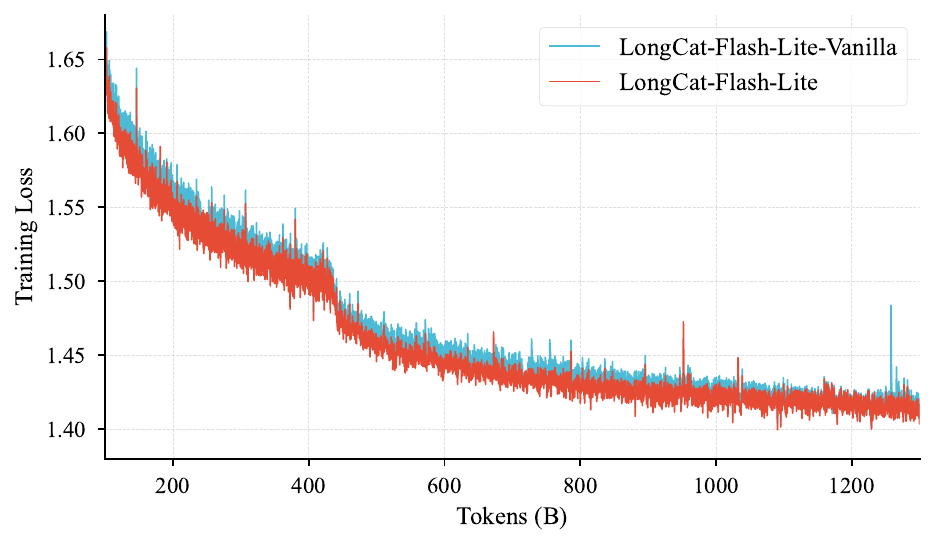}
    \caption{Smoothed training loss curves of \longcat and \longcat-Vanilla. The loss drop at 420B tokens coincides with the batch size increases.}
    \label{fig:lm-loss}
\end{figure}

Throughout training, \longcat consistently achieves lower training loss compared to \longcat-Vanilla, as illustrated in Figure~\ref{fig:lm-loss}. To assess downstream performance, we evaluate both models on benchmarks spanning three core capability domains:
\begin{itemize}
    \item \textbf{General Tasks:} MMLU~\citep{hendrycks2021measuringmassivemultitasklanguage}, MMLU-Pro~\citep{wang2024mmluprorobustchallengingmultitask}, C-Eval~\citep{huang2023ceval}, and CMMLU~\citep{li2023cmmlu}.
    \item \textbf{Reasoning Tasks:} BBH~\citep{BBH}, GPQA~\citep{pteam2025supergpqascalingllmevaluation}, DROP~\citep{drop} and GSM8K~\citep{cobbe2021gsm8k}.
    \item \textbf{Coding Tasks:} %
    HumanEval+~\citep{humanevalmbppplus}, MultiPL-E~\citep{cassano2022multiplescalableextensibleapproach}, and BigCodeBench~\citep{zhuo2025bigcodebenchbenchmarkingcodegeneration}. 
\end{itemize}

\begin{table}[htbp]
    \centering
    \caption{Comparison of base model between \longcat and \longcat-Vanilla.}
    \resizebox{0.8\textwidth}{!}{
    \begin{tabular}{ll|c|c}
        \toprule
        ~ & BenchMark & \longcat-Vanilla@1.3T & \longcat@1.3T \\
        \midrule
        \multirow{4}{*}{General} & MMLU & \textbf{64.81} & 64.01 \\
        & MMLU-pro & 34.43 & \textbf{35.89} \\
        & CEval & 64.09 & \textbf{67.21} \\
        & CMMLU & 67.08 & \textbf{69.55} \\
        \midrule
        \multirow{4}{*}{Reasoning} & BBH & 38.54 & \textbf{43.67} \\
        & GPQA & 25.37 & \textbf{29.66} \\
        & DROP & 47.92 & \textbf{52.43} \\
        & GSM8K & 50.00 & \textbf{50.50} \\
        \midrule
        \multirow{3}{*}{Coding} %
        & HumanEval+ & 28.66 & \textbf{31.10} \\
        & MultiPL-E & \textbf{30.20} & 30.03 \\
        & BigCodeBench & 33.42 & \textbf{36.05} \\
        \bottomrule
    \end{tabular}
    }
    \label{tab:base_model_results}
\end{table}

As detailed in Table~\ref{tab:base_model_results}, \longcat demonstrates substantial performance improvements over \longcat-Vanilla across the majority of benchmarks in all three domains. These findings validate our earlier analysis: when sparsity reaches sufficient levels, strategically scaling total parameters through \method —while maintaining an optimal proportion of embedding parameters—consistently outperforms approaches that merely increase expert numbers.

\subsection{Chat Model Evaluation} 
The evaluation of the chat model covers several core capabilities: agentic tool use tasks, agentic coding tasks, general domain tasks and mathematical reasoning tasks. The benchmarks used for assessment include:
\begin{itemize}
    \item \textbf{Agentic Tool Use Tasks:} $\tau^2$ Bench~\citep{tau2-bench}, Vita Bench ~\citep{he2025vitabench}. For $\tau^2$-Bench, we use a revised version\footnote{\url{https://github.com/AGI-Eval-Official/tau2-bench-revised}} to perform our evaluation because the original version contains noisy data.
    \item \textbf{Agentic Coding Tasks:} SWE-Bench ~\citep{jimenez2023swe}, 
    TerminalBench ~\citep{merrill2026terminalbenchbenchmarkingagentshard}, SWE-Bench Multiligual~\citep{yang2025swesmith}, and PRDBench~\citep{fu2025automatically}. 
    \item \textbf{General Domain Tasks:} GPQA-Diamond ~\citep{rein2024gpqa}, MMLU~\citep{hendrycks2021measuringmassivemultitasklanguage}, MMLU-Pro~\citep{wang2024mmluprorobustchallengingmultitask}, C-Eval~\citep{huang2023ceval}, and CMMLU~\citep{li2023cmmlu}.
    \item \textbf{Mathematical Reasoning Tasks:} MATH500~\citep{math500}, AIME24~\citep{AIME24}, AIME25~\citep{AIME25}.
\end{itemize}

\begin{table}[htbp]
    \caption{Comparison between \longcat and other models. Values marked with * are sourced from public reports.}
    \centering
    \resizebox{\textwidth}{!}{
    \begin{tabular}{l|ccc|c}
        \toprule
        Benchmark & \begin{tabular}[c]{@{}c@{}} Kimi-Linear-48B-A3B \end{tabular} & \begin{tabular}[c]{@{}c@{}} Qwen3-Next-80B-A3B-Instruct \end{tabular} & \begin{tabular}[c]{@{}c@{}} Gemini 2.5 Flash-Lite\end{tabular}  & \begin{tabular}[c]{@{}c@{}}\longcat \end{tabular} \\ 
        \midrule
        Architecture & MoE & MoE  & -  & MoE + NE\\
        \# Total Params & 48B & 80B &  - & 68.5B\\
        \# Activated Params & 3B & 3B &  - & 2.9B$\sim$4.5B\\
        \midrule
        \multicolumn{5}{c}{\textbf{Agentic Tool Use}} \\
        \midrule
        Tau2-Airline(avg@8) & 44.00 & 45.5* & 35.00 & {58.00} \\
        Tau2-Retail(avg@8) & 18.86 & 57.3* & 37.50 & {73.10} \\
        Tau2-Telecom(avg@8) & 15.68 & 13.2* & 21.93 & {72.80} \\
        VitaBench(avg@4) & - & 5.80 & 4.50  & {7.00}\\
        \midrule
        \multicolumn{5}{c}{\textbf{Agentic Coding}} \\
        \midrule
        SWE-Bench(acc) & 32.80 & 37.60 & 41.3* & {54.40}\\
        TerminalBench(acc) & 20.00 & 15.19 & 20.00 & {33.75}\\
        SWE-Bench Multiligual & 37.20 & 31.30 & - &  {38.10}\\
        PRDBench & - & 15.36 & - &  {39.63}\\
        \midrule
        \multicolumn{5}{c}{\textbf{General Domains}} \\
        \midrule
        GPQA-Diamond(avg@16) & 69.89 & {74.33} & 70.20* & 66.78\\
        MMLU(acc) & 79.91 & {89.28} & 84.68 & 85.52\\
        MMLU-Pro(acc) & 67.22 & {82.93} & 78.95 & 78.29\\
        CEval(acc) & 78.48 & {90.91} & 75.16  & 86.55 \\
        CMMLU(acc) & 76.26 & {86.50} & 72.06  & 82.48 \\
        \midrule
        \multicolumn{5}{c}{\textbf{Mathematical Reasoning}} \\
        \midrule
        MATH500(acc) & 94.20 & {98.00} & 95.20  & 96.80\\
        AIME24(avg@32) & 70.52 & {81.35} & 63.33 & 72.19 \\
        AIME25(avg@32) & 59.58 & {68.44} & 50.1*  & 63.23 \\
        \bottomrule
    \end{tabular}
    }
    \label{tab:chat_model_results}
\end{table}

Table \ref{tab:chat_model_results} presents the comprehensive evaluation results of \longcat across various benchmark categories, along with comparisons with Qwen3-Next-80B-A3B-Instruct, Gemini 2.5 Flash-Lite\footnote{Gemini 2.5 Flash-Lite Preview 09-2025}, and Kimi-Linear-48B-A3B. %
\longcat demonstrates exceptional parameter efficiency and competitive performance across core capability dimensions.

\textbf{Agentic Tool Use.} \longcat excels in agentic tool use tasks, establishing a clear lead over all comparison models. In the $\tau^2$-Bench benchmark, it achieves the highest scores across all three sub-scenarios: Telecom (72.8), Retail (73.1), and Airline (58.0). Notably, in the Telecom scenario, its score significantly outperforms Gemini 2.5 Flash-Lite and Kimi-Linear-48B-A3B. This highlights its superior ability to handle complex dependencies on tools and domain-specific task execution. In VitaBench, it achieves a score of 7.00, outperforming Qwen3-Next-80B-A3B-Instruct (5.80), and Gemini 2.5 Flash-Lite (4.50). This leading score underscores \longcat's superior ability to handle complex, real-world task workflows via tool integration in practical business scenarios.

\textbf{Agentic Coding.} In coding-related tasks, \longcat demonstrates remarkable practical problem-solving capabilities. In SWE-Bench, it achieves an accuracy of 54.4, outperforming all baselines—surpassing Qwen3-Next-80B-A3B-Instruct (37.6), Gemini 2.5 Flash-Lite (41.3), and Kimi-Linear-48B-A3B (32.8) by a significant margin. This indicates its proficiency in solving real-world software engineering issues, including bug fixes and feature implementation. In TerminalBench, which evaluates terminal command execution competence, \longcat secures a leading score of 33.75, far exceeding Qwen3-Next-80B-A3B-Instruct (15.19), Gemini 2.5 Flash-Lite (20.0) and Kimi-Linear-48B-A3B (20.0), reflecting its robust ability to understand and execute terminal-related instructions critical for developer-centric agentic applications. Additionally, in SWE-Bench Multilingual—a benchmark designed to measure cross-language programming generalization across diverse software ecosystems—\longcat achieves a strong accuracy of 38.10, outperforming Qwen3-Next-80B-A3B-Instruct (31.3) and Kimi-Linear-48B-A3B (37.2), thus demonstrating its reliable adaptability to multi-language development scenarios. In PRDBench, \longcat achieves a score of 39.63, significantly outperforming Qwen3-Next-80B-A3B-Instruct (15.36). We observe that our model can autonomously write unit tests to verify its development, producing higher-quality code repositories.

\textbf{General Domains.} \longcat delivers balanced and competitive performance in general domain knowledge tasks. On MMLU, it scores 85.52, which is comparable to Gemini 2.5 Flash-Lite (84.68) and Kimi-Linear-48B-A3B (79.91), and only slightly lower than Qwen3-Next-80B-A3B-Instruct (89.28). In Chinese-specific benchmarks (CEval and CMMLU), it achieves 86.55 and 82.48 respectively, performing particularly well against Kimi-Linear-48B-A3B (78.48 and 76.26) and Gemini 2.5 Flash-Lite (75.16 and 72.06). On GPQA-Diamond, it scores 66.78, maintaining competitiveness within the benchmark’s performance range. For MMLU-Pro, it achieves 78.29, demonstrating solid performance in handling more challenging multi-task language understanding questions.

\textbf{Mathematical Reasoning.} \longcat exhibits strong mathematical reasoning capabilities across both basic and advanced tasks. On MATH500, it achieves an accuracy of 96.80, which is close to Qwen3-Next-80B-A3B-Instruct (98.00), and outperforms Gemini 2.5 Flash-Lite (95.20). In advanced mathematical competition benchmarks, it delivers impressive results: AIME24 (72.19) and AIME25 (63.23). These scores surpass Kimi-Linear-48B-A3B (70.52 and 59.58) and Gemini 2.5 Flash-Lite (63.33 and 50.1), highlighting its ability to handle complex, multi-step mathematical deduction.

\subsection{Fast Inference with Optimized Kernels}

As discussed in Section~\ref{sec:inference}, the extreme activation sparsity of this model necessitates a large effective batch size to fully saturate GPU memory bandwidth. To achieve this, we deploy the model using ``Eagle3'' ~\citep{li2025eagle3scalinginferenceacceleration} with a ``3-step speculative decoding strategy''. Similarly to \cite{qian2025epsmoeexpertpipelinescheduler} and \cite{meituanlongcatteam2025longcatflashtechnicalreport}, we adopt wide EP~(Expert Parallel) and SBO~(Single Batch Overlap) to accelerate inference speed. While the above optimizations successfully expands the effective batch size, the model's lightweight nature shifts the bottleneck towards kernel launch overheads, making it challenging to maintain high GPU occupancy. To address this and minimize end-to-end latency, we implement the following system-level optimizations:

\paragraph{Kernel Optimization}
    \begin{itemize}
        \item Kernel Fusion: We apply extensive kernel fusion to reduce execution overhead and memory traffic. Specifically, all intra-TP-group communication operations are fused with subsequent fine-grained kernels (e.g., \texttt{AllReduce} + \texttt{Residual Add} + \texttt{RMSNorm}, \texttt{AllGather} + \texttt{Q-Norm} + \texttt{KV-Norm}, and \texttt{ReduceScatter} + \texttt{RMSNorm} + \texttt{Hidden State Combine}). For the quantized model, we integrate every activation quantization step into existing operators, including the aforementioned communication-fusion kernels and the \texttt{SwiGLU} component. Additionally, the processing of router logits (\texttt{Softmax} + \texttt{TopK} + \texttt{router scaling}) and zero-expert selection is consolidated into a single unified kernel.
        \item Optimized Attention Combine: We employ a splitkv-and-combine strategy during decoding phase. When the number of KV splits is high, the combine operation can incur significant latency, sometimes comparable to the computation itself. By optimizing the combine kernel, we effectively reduce its latency by 50\%.
    \end{itemize}

\paragraph{PDL (Programmatic Dependent Launch)} 
We utilize PDL ~\citep{pdl_nvidia} to allow dependent kernels to overlap their execution by triggering early launches. This mechanism not only eliminates the gaps between consecutive kernels but also improves SM utilization.

Building upon these optimizations, we achieve the exceptional inference performance illustrated in Figure~\ref{fig:decode-performance}.

\section{Conclusions}
In this technical report, we presented a comprehensive study on the scalability and efficiency of embedding scaling in LLMs. Through systematic analysis of architectural constraints and comparative scaling laws, we demonstrated that scaling embeddings yields a superior Pareto frontier compared to increasing expert numbers in specific regimes, while our proposed system optimizations, including the N-gram Cache and synchronized kernels, effectively resolve associated I/O bottlenecks. Validating these findings, we introduced \longcat, a 68.5B MoE model with over 30B \method parameters, which not only outperforms parameter-equivalent MoE baselines but also exhibits competitive performance in agentic and coding tasks, thereby establishing a robust and efficient framework for future model scaling.

\section{Acknowledgement}
We extend our sincere gratitude to both the infrastructure team and evaluation team for their invaluable support and constructive feedback throughout this project. The primary contributors from these teams include:

\begin{tabular}{p{0.25\textwidth}p{0.25\textwidth}p{0.25\textwidth}p{0.25\textwidth}}
Linsen Guo & Lin Qiu & Xiao Liu & Yaoming Zhu \\
Mengxia Shen & Zijian Zhang & Xiaoyu Li & Chao Zhang \\
Yunke Zhao & Dengchang Zhao & Yifan Lu &  \\
\end{tabular}

\clearpage

\bibliographystyle{unsrtnat}
\bibliography{references}

\end{document}